\documentclass{tlp}
\usepackage{amsmath}

\newcommand{\seq}[1]{\textit{seq}(#1)}

\DeclareMathOperator*{\bigcomma}{\textrm{\huge ,}}
\DeclareMathOperator*{\bigscolon}{\textrm{\huge ;}}

\newtheorem{theorem}{Theorem}
\newtheorem{theorem2}{Theorem}
\newtheorem{proposition}{Proposition}
\newtheorem{fact}{Fact}
\newtheorem{lemma}{Lemma}

\def\ii#1{\hbox{\it #1\/}}
\def\is#1{\hbox{\scriptsize\it #1\/}}
\def\no{\ii{not}}
\def\sno{\is{not}}

\def\ar{\leftarrow}

\title{Weight Constraints as Nested Expressions%
}

\author[P. Ferraris and V. Lifschitz]{
PAOLO FERRARIS and VLADIMIR LIFSCHITZ\\
Department of Computer Sciences\\
University of Texas at Austin\\
Austin, TX 78712, USA\\
\email{\{otto,vl\}@cs.utexas.edu}
}

\date{}
\begin{document}
\bibliographystyle{acmtrans}

\maketitle

\begin{abstract}
We compare two recent extensions of the answer set (stable model)
semantics of logic programs.
One of them, due to Lifschitz, Tang and Turner, allows the
bodies and heads of rules to contain nested expressions.  The other,
due to Niemel\"a and Simons, uses weight constraints.  We show that
there is a simple, modular translation from the language of weight
constraints into the language of nested expressions that preserves the
program's answer sets.
Nested expressions can be eliminated from the result of this translation
in favor of additional atoms.
The translation makes it possible to compute answer sets for some programs
with weight constraints using satisfiability solvers, and to prove the
strong equivalence of programs with weight constraints using the logic of
here-and-there.

\end{abstract}

\section{Introduction}

The notion of an answer set (or ``stable model'') was defined in
\cite{gel88} for logic programs whose rules have simple syntactic
structure.
The head of such a rule is an atom.  The body is a list of atoms, some
of them possibly prefixed with the negation as failure symbol.  In this
paper, we compare two recent extensions of the answer set semantics.

In one of these extensions, the head and the body of a rule are allowed to
contain negation as failure (\no), conjunction (,) and disjunction (;),
nested arbitrarily~\cite{lif99d}.  In particular, negation as failure can
occur in the head of a rule, as proposed in \cite{lif92b}.  For instance,
\begin{equation}
\label{ex2rule}
a;\no\ a
\end{equation}
is a rule with the empty body.  The program  whose only rule is~(\ref{ex2rule})
can be shown to have two answer sets: $\emptyset$ and  $\{a\}$.  The rule
\begin{equation}
\label{ex2arule}
a\ar \no\ \no\ a
\end{equation}
is another example of a rule with nested expressions.  It is
``nondisjunctive''---its head is an atom; but its body contains nested
occurrences of negation as failure.  The program whose only rule
is~(\ref{ex2arule}) has the same answer sets as~(\ref{ex2rule}).

According to the second proposal~\cite{nie00},  rules are allowed to
contain ``cardinality constraints'' and more general ``weight
constraints.''
For instance,
\begin{equation}
\label{ex1arule}
0\leq \{a,b\}\leq 1
\end{equation}
is a cardinality constraint.  This expression can be viewed as a logic
program consisting of a single rule
with the empty body.  Its answer sets are the subsets of
$\{a,b\}$ whose cardinalities are between 0 and 1, that is to say, sets
$\emptyset$, $\{a\}$ and $\{b\}$.

Cardinality and weight constraints are important elements of the input
language of {\sc smodels} --- a software system for computing answer
sets that can be used to solve many kinds of combinatorial search
problems.\footnote{\tt http://www.tcs.hut.fi/Software/smodels/ .}
The idea of this programming method, called {\sl answer set programming},
is to represent the given search problem by a logic program whose answer
sets correspond to solutions.   Cardinality constraints are found in many
programs of this kind.

It may appear that the two extensions of the basic syntax of logic
programs --- nested expressions and weight constraints --- have little in
common.  The following observation suggests that it would
not be surprising actually if these ideas were related to each other.
The original definition of
an answer set is known to have the ``anti-chain'' property: an answer
set for
a program cannot be a subset of another answer set for the same program.
Examples~(\ref{ex2rule}) and~(\ref{ex2arule}) show that the anti-chain
property is lost as soon as nested expressions are allowed in rules.
Example~(\ref{ex1arule}) shows
that in the presence of cardinality constraints the anti-chain property does
not hold either.

In this paper we show that there is indeed a close relationship between
these two forms of the answer set semantics: cardinality and
weight constraints can be viewed as shorthand for nested expressions of
a special form.  We define a simple, modular translation that turns any
program $\Omega$
with weight constraints into a program $[\Omega]$ with nested expressions
that has the same answer sets as $\Omega$.   Furthermore,
every rule of $[\Omega]$ can be equivalently replaced with a set of
nondisjunctive rules, and this will lead us to a nondisjunctive
version~$[\Omega]^{nd}$ of the basic translation.  Finally, we will define
a ``nonnested translation''~$[\Omega]^{nn}$, obtained from~$[\Omega]^{nd}$
by eliminating nested expressions in the bodies of rules in favor of
additional atoms.  The 
nonnested translation is a conservative extension of $\Omega$, in the
sense that dropping the new atoms from its answer sets gives the answer
sets for~$\Omega$.

The translations defined in this paper can be of interest for several
reasons.  First, the definition of an answer set for programs with weight
constraints is technically somewhat complicated.  Instead of introducing that
definition, we can treat any program $\Omega$ with weight constraints as
shorthand for its translation~$[\Omega]$.

Second, the definition of program completion from~\cite{cla78} has been
extended to nondisjunctive programs with nested expressions~\cite{llo84a},
and this extension is known to be equivalent to the definition of an answer
set whenever the program is ``tight''~\cite{erd03}.  In view of this fact,
answer sets for a tight logic program
can be generated by running a satisfiability solver on the program's
completion~\cite{bab00}.  Consequently, answer sets for a program~$\Omega$
with weight constraints can be computed by running a satisfiability solver
on the completion of one of the translations~$[\Omega]^{nd}$,~$[\Omega]^{nn}$,
if that translation is tight.  This idea has led to the creation of a new
software system for computing answer sets, called
{\sc cmodels}\footnote{\tt http://www.cs.utexas.edu/users/tag/cmodels.html .};
see \cite[Section 7]{erd03} for details.

Third, recent work on the theory of logic programs with nested expressions
has led to a simple theory of equivalent transformations of such
programs.  Two programs are said to be {\sl weakly equivalent} if they have the
same answer sets, and {\sl strongly equivalent} if they remain weakly
equivalent after adding an arbitrary set of rules to both of them.  For
instance, rule~(\ref{ex2arule}) is strongly equivalent to rule~(\ref{ex2rule}),
so that replacing one of these rules by the other in any program does not
affect that program's answer sets.
The study of strong equivalence is important because we learn from it how
one can simplify a part of a program without looking at the other parts.
The main theorem of~\cite{lif01} shows
that the strong equivalence of programs with nested expressions is
characterized by the truth tables of the three-valued logic known as the
logic of here-and-there.\footnote{The close relationship between answer sets
and the logic of here-and-there was first discovered by
Pearce~[\citeyear{pea97}].}
Our translations can be used to prove
the strong equivalence of programs with weight constraints using that logic.

The possibility of translating programs with cardinality constraints into
the language of nonnested programs at the price of introducing
new atoms was first established by Marek and Remmel~[\citeyear{mar02}].  Our
nonnested translation is more general, because it is applicable to programs
with arbitrary weight constraints.  Its other advantage is that, in the
special case when all weights in the program are expressed by integers of a
limited size (in particular, in the case of cardinality constraints) it does
not make the program exponentially bigger.  (In the translation from
\cite[Section 3]{mar02}, the number of rules introduced in part (II) can be
exponentially large.)\footnote{The
use of additional atoms to keep the program small in the process of
eliminating nested expressions is discussed in~\cite{pea02}.  In case of
the transition from~$[\Omega]^{nd}$ to~$[\Omega]^{nn}$, the role of 
additional atoms is even more significant: both the basic and nondisjunctive 
translations can be exponentially bigger than the original program, and the
use of new atoms allows us
to scale~$[\Omega]^{nd}$ back down approximately to the size of $\Omega$.
The other reason why we are not applying here the translation
from \cite{pea02} to~$[\Omega]^{nd}$ is that it would
not give a nondisjunctive program.}

We begin with a review of programs with nested
expressions~(Section~\ref{sec:nested}) and programs with weight
constraints~(Section~\ref{sec:weight}).  The translations are defined in
Section~\ref{sec:translation}, and their use for proving strong equivalence
of programs with weight constraints is discussed in
Section~\ref{sec:applications}.  Proofs of more difficult theorems are
relegated to Section~\ref{sec:proof}.  Some properties of programs with
nested expressions proved in that section, such as the completion lemma and
the lemma on explicit definitions, may be of more general interest.

\section{Programs with Nested Expressions}\label{sec:nested}

\subsection{Syntax}\label{sec:syntaxnested}

A {\sl literal} is a propositional atom possibly prefixed with the classical
negation sign~$\neg$.  {\sl Elementary formulas} are literals and the
symbols
$\bot$ (``false'') and $\top$ (``true''). {\sl Formulas} are built from
elementary formulas
using the unary connective $\no$ (negation as failure)
and the binary connectives $,$
(conjunction) and $;$ (disjunction).
A {\sl rule with nested expressions} has the form
\begin{equation}
\label{genrule}
\ii{Head} \ar \ii{Body}
\end{equation}
where both $\ii{Body}$ and $\ii{Head}$ are formulas.  For
instance,~(\ref{ex2rule}) is a formula; it can be used as shorthand
for the rule
$$a;\no\ a\ar\top.$$
The expression
\begin{equation}
\label{neglit}
\neg a \ar \no\ a
\end{equation}
is an example of a rule containing classical negation.

A {\sl program with nested expressions} is any set of rules with
nested expressions.

\subsection{Semantics}\label{sec:semanticsnested}

The semantics of programs with nested expressions is characterized by
defining
when a consistent set $Z$ of literals is an answer set for a program $\Pi$.
As a preliminary step, we define when a consistent set $Z$ of literals
{\sl satisfies} a formula $F$ (symbolically, $Z \models F$), as follows:
\begin{itemize}
\item for a literal $l$, $Z\models l$ if $l\in Z$
\item $Z\models \top$
\item $Z\not\models \bot$
\item $Z\models (F,G)$ if $Z\models F$ and $Z\models G$
\item $Z\models (F;G)$ if $Z\models F$ or $Z\models G$
\item $Z\models \no\ F$ if $Z\not\models F$.
\end{itemize}
We say that $Z$ {\sl satisfies} a program $\Pi$ (symbolically, $Z
\models \Pi$)
if, for every rule~(\ref{genrule}) in~$\Pi$, $Z \models \ii{Head}$ whenever
$Z \models~\ii{Body}$.

The {\sl reduct}\footnote{This definition of reduct is the same as the one
in~\cite{lif01}, except that the condition $Z\models F^Z$ is replaced
with $Z\models F$. It is easy to check by structural induction that the two
conditions are equivalent.} $F^Z$ of a formula $F$ with respect to a
consistent set $Z$ of literals is defined recursively as follows:
\begin{itemize}
\item
    for elementary $F$, $F^Z=F$
\item
    $(F,G)^Z=F^Z,G^Z$
\item
    $(F;G)^Z=F^Z;G^Z$
\item
    $(\no\ F)^Z=
\begin{cases}
\bot\ , & \text{if $Z \models F$}, \cr
\top\ , & \text{otherwise}\hfill \end{cases}$

\end{itemize}
The {\sl reduct} $\Pi^Z$ of a program $\Pi$ with respect to $Z$
is the set of rules
$$
\ii{Head}^Z \ar \ii{Body}^Z
$$
for each rule~(\ref{genrule}) in $\Pi$.
For instance, the reduct of (\ref{ex2arule}) with respect
to $Z$ is
\begin{equation}
a\ar\top
\label{red1}
\end{equation}
if $a\in Z$,
and
\begin{equation}
a\ar\bot
\label{red2}
\end{equation}
otherwise.

The concept of an answer set is defined first for programs not containing
negation as failure: a consistent set $Z$ of literals is an {\sl answer set}
for such a program $\Pi$ if $Z$ is a minimal set satisfying $\Pi$.  For an
arbitrary program $\Pi$, we say that $Z$ is an {\sl answer set} for $\Pi$ if
$Z$ is an answer set for the reduct $\Pi^Z$.

For instance, the reduct
of~(\ref{ex2arule}) with respect to $\{a\}$ is~(\ref{red1}), and $\{a\}$ is
a minimal set satisfying~(\ref{red1}); consequently, $\{a\}$ is an
answer set
for~(\ref{ex2arule}).  On the other hand, the reduct of~(\ref{ex2arule})
with respect to $\emptyset$ is~(\ref{red2}), and $\emptyset$ is
a minimal set satisfying~(\ref{red2}); consequently, $\emptyset$ is an
answer set for~(\ref{ex2arule}) as well.

\subsection{A Useful Abbreviation}\label{sec:abbr}

The following abbreviation is used in the definition of the translation
$[\Omega]$ in Section~\ref{sec:translation}.  For
any formulas $F_1,\dots,F_n$ and any set $X$ of subsets of $\{1,\dots,n\}$, by
$$
\langle F_1,\dots,F_n\rangle:X
$$
we denote the formula
\begin{equation}
\label{inlargeset}
\bigscolon_{I \in X}
\big( \bigcomma_{i \in I} F_i\big).
\end{equation}
The use of the ``big comma'' and the ``big semicolon'' in (\ref{inlargeset})
to represent a multiple conjunction and a multiple disjunction is similar
to the familiar use of $\bigwedge$ and $\bigvee$.  In particular, the empty
conjunction is understood as~$\top$, and the empty disjunction as~$\bot$.

For instance, if $X$ is the set of all subsets of $\{1,\dots,n\}$ of
cardinality $\geq 3$, then~(\ref{inlargeset}) expresses, intuitively,
that at least 3 of the formulas $F_1,\dots,F_n$ are true.  It is easy to
check, for this~$X$, that a consistent set $Z$ of literals
satisfies~(\ref{inlargeset}) iff $Z$ satisfies at least 3 of the
formulas~$F_1,\dots,F_n$.  This observation can be generalized:

\begin{proposition}
\label{prop1}
Assume that for every subset $I$ of $\{1,\dots,n\}$ that belongs to $X$,
all supersets of $I$ belong to $X$ also. For any formulas
$F_1,\dots,F_n$ and
any consistent set $Z$ of literals,
$$
Z\models \langle F_1,\dots,F_n\rangle:X\hbox{ iff }\{i:Z\models F_i\} \in X.
$$
\end{proposition}

{\samepage
\noindent
\begin{proof*}
$$
\begin{array}{rcl}
Z\models \langle F_1,\dots,F_n \rangle:X &\hbox{iff}&
    \hbox{ for some $I\in X$, for all $i$, if $i\in I$ then $Z\models
F_i$}\\
     &\hbox{iff}& \hbox{ for some $I\in X$, $I\subseteq \{i: Z\models
F_i$\}}\\
     &\hbox{iff}&
    \hbox{ for some $I\in X$, $I= \{i: Z\models F_i$\}}\\
     &\hbox{iff}&
    \{i: Z\models F_i\}\in X. \mathproofbox
\end{array}
$$
\end{proof*}
}

\subsection{Strong Equivalence}\label{sec:ht}

Recall that programs $\Pi_1$ and $\Pi_2$ are said to be {\sl strongly
equivalent} to each other
if, for every program $\Pi$, the union $\Pi_1\cup \Pi$ has the
same answer sets as  $\Pi_2\cup \Pi$.
This concept is essential both for applications of our translations
and for the proof of their soundness.

The method of proving strong equivalence proposed in~\cite{lif01} is
particularly simple for programs that do not contain classical negation.
We first rewrite both programs in the syntax of propositional formulas by
writing every rule~(\ref{genrule}) as the implication
$\ii{Body}\rightarrow \ii{Head}$ and replacing every comma in the rule
with~$\wedge$, every semicolon with~$\vee$, and every occurrence of
negation as failure with~$\neg$.  For instance, rule~(\ref{ex2arule}) in
this notation is
$$\neg\neg a \rightarrow a.$$
Then we check whether the rules of each of the programs $\Pi_1$, $\Pi_2$
are entailed by the rules of the other in the logic of here-and-there; if
they are, then $\Pi_1$ and $\Pi_2$ are strongly equivalent to each other, and
the other way around (\cite{lif01}, Theorem 1).

The logic of here-and-there was originally defined in~\cite{hey30}.  Its
definition and basic properties are discussed in~\cite[Section~2]{lif01}.
It is a three-valued logic, intermediate between intuitionistic and classical.
Recall that a natural deduction system for
intuitionistic logic can be obtained from the corresponding classical
system~\cite[Table~3]{bib93} by dropping the law of the excluded middle
$$F\vee\neg F$$
from the list of postulates. The logic of here-and-there, on the other hand,
is the result of replacing the excluded middle in the classical system with
the weaker axiom schema
$$
F\vee(F\rightarrow G)\vee\neg G.
$$
In addition to all intuitionistically provable formulas, the set of
theorems of the logic of here-and-there includes, for
instance, the weak law of the excluded middle
$$
\neg F\vee\neg\neg F
$$
and De Morgan's law
$$
\neg(F\wedge G)\leftrightarrow \neg F \vee \neg G
$$
(the dual law can be proved even intuitionistically).

As an example of the use of this idea, note that the absorption laws
$$
\begin{array}c
a \vee (a \wedge b)\leftrightarrow a\\
a \wedge (a \vee b)\leftrightarrow a\\
\end{array}
$$
are provable in the logic of here-and-there (their usual proofs in the
natural
deduction formalization of propositional logic do not use the law of the
excluded middle).  It follows that, in any program, $a;(a,b)$ and $a,(a;b)$
can be replaced by $a$ without changing the program's answer sets.
In particular, if we take a program containing a
multiple disjunction of the form~(\ref{inlargeset}) and restrict this
disjunction to the sets $I$ that are minimal in $X$, then the answer sets of
the program will remain the same.

As another example, let us verify that~(\ref{ex2rule}) is strongly
equivalent
to~(\ref{ex2arule}) by proving the equivalence
$$a\vee\neg\ a \leftrightarrow \neg\neg a \rightarrow a$$
in the logic of here-and-there.  The proof left-to-right
is straightforward, by considering the cases $a$, $\neg a$.
Right-to-left,
use the instance $\neg a \vee \neg\neg a$ of the weak law of the
excluded
middle and consider the cases $\neg a$, $\neg \neg a$.

The extension of this method to programs with classical negation is based
on the fact that classical negation can be eliminated from any program~$\Pi$
by a simple syntactic transformation.  For every atom $a$ that occurs in $\Pi$
after the classical negation symbol~$\neg$, choose a new atom $a'$ and replace
all occurrences of $\neg a$ with~$a'$.  The answer sets for the resulting
program~$\Pi'$ that do not contain any of the pairs $\{a,a'\}$
are in a 1--1 correspondence with the answer sets
for~$\Pi$~\cite[Section~5]{lif01}.  If the rules of each of the programs
$\Pi_1'$, $\Pi_2'$ can be derived from the rules of the other program and the
formulas $\neg(a\wedge a')$ in the logic of here-and-there then $\Pi_1$ and
$\Pi_2$ are strongly equivalent to each other, and the other way around
(\cite{lif01}, Theorem 2).

\section{Programs with Weight Constraints}\label{sec:weight}

\subsection{Syntax}\label{sec:syntaxweight}

A {\sl rule element} is a literal ({\sl positive} rule element) or a
literal prefixed with $\no$ ({\sl negative} rule element).
A {\sl weight constraint} is an expression of the form
\begin{equation}
L\leq \{ c_1 = w_1, \ldots, c_m = w_m\}\leq U
\label{wc}
\end{equation}
where
\begin{itemize}
\item each of $L$, $U$ is (a symbol for) a real number or one
of the symbols $-\infty$, $+\infty$,
\item $c_1,\dots,c_m$ ($m\geq 0$) are rule elements, and
\item $w_1,\dots,w_m$ are nonnegative real numbers
(``weights'').\footnote{\samepage In~\cite{sim02}, weights are not required to
be nonnegative, and the meaning of a program with negative weights is defined
by describing
a method for eliminating them.  Unfortunately, this preprocessing step
leads to some results that seem unintuitive.  For instance, it turns
out that the one-rule programs
$$
1\leq\{p=1\}\leftarrow 0\leq\{p=2,p=-1\}
$$
and
$$
1\leq\{p=1\}\leftarrow 0\leq\{p=1\}
$$
have different answer sets.
}
\end{itemize}
The part $\ L\leq\ $ can be omitted if $L = -\infty$; the part
$\ \leq U\ $ can be omitted if $U = +\infty$.
A {\sl rule with weight constraints} is an expression of the form
\begin{equation}
\label{rule}
C_0 \ar C_1, \ldots, C_n
\end{equation}
where $C_0,\dots,C_n$ ($n\geq 0$) are weight constraints.
We will call the rule elements of $C_0$ the
{\sl head elements} of rule~(\ref{rule}).

Finally, a {\sl program with weight constraints} is a set of rules with
weight constraints.\footnote{In~\cite{sim02}, programs are not allowed to
contain classical negation.  But classical negation is allowed in the input
files of the current version of {\sc smodels}.}

This syntax becomes a generalization of the basic syntax of logic
programs for which the answer set semantics was originally
defined~\cite{gel88} if we identify a rule element $c$ with the
weight constraint
$$
1\leq \{c=1\}.
$$
By
$$\ar C_1, \ldots, C_n$$
we denote the rule
$$
1\leq \{\ \} \ar C_1, \ldots, C_n.
$$

A {\sl cardinality constraint} is a weight constraint with all weights
equal to 1.  A cardinality constraint
$$L\leq \{ c_1 = 1, \ldots, c_m = 1\}\leq U$$
can be abbreviated as
\begin{equation}
L\leq \{ c_1, \ldots, c_m\}\leq U.
\label{cc}
\end{equation}

\subsection{Semantics}\label{sec:semanticsweight}

The definition of an answer set for programs with weight constraints
in~\cite{sim02} uses the following auxiliary definitions.
A consistent set $Z$ of literals {\sl satisfies} a weight
constraint~(\ref{wc}) if the sum of the weights $w_j$ for all $j$ such
that $Z \models c_j$ is not less
than $L$ and not greater than $U$.  For instance, $Z$ satisfies cardinality
constraint~(\ref{ex1arule}) iff $Z$ contains at most one of the atoms
$a$, $b$.
About a program $\Omega$
with weight constraints we say that $Z$ {\sl satisfies} $\Omega$ if, for
every rule~(\ref{rule}) in~$\Omega$, $Z$ satisfies~$C_0$ whenever $Z$
satisfies
$C_1,\dots,C_n$.  As in the
case of nested expressions, we will use $\models$ to
denote the satisfaction relation for both weight constraints and programs
with weight constraints.

The next part of the semantics of weight constraints is the definition of
the reduct for weight constraints of the form
$$L\leq \{ c_1 = w_1, \ldots, c_m = w_m\}.$$
The {\sl reduct} $(L\leq S)^Z$ of a weight constraint $L\leq S$
with respect to a consistent set $Z$ of literals is the weight constraint
$L^Z\leq S'$, where
\begin{itemize}
\item $S'$ is obtained from $S$ by dropping all pairs $c = w$ such that
$c$
is negative, and
\item $L^Z$ is $L$ minus the sum of the weights $w$ for all pairs $c = w$
in $S$ such that $c$ is negative and $Z\models c$.
\end{itemize}
For instance, the reduct of the constraint
$$1\leq\{\no\ a=3,\no\ b=2\}$$
relative to $\{a\}$ is
$$-1\leq\{\ \}.$$

The {\sl reduct} of a rule
\begin{equation}
\label{weightrule}
L_0\leq S_0\leq U_0 \ar L_1\leq S_1\leq U_1,\ldots,L_n\leq S_n\leq U_n
\end{equation}
with respect to a consistent set $Z$ of literals is
\begin{itemize}
\item the set of rules of the form
$$
l \ar (L_1\leq S_1)^Z,\ldots,(L_n\leq S_n)^Z
$$
where $l$ is a positive head element of~(\ref{weightrule}) such that
$Z \models l$,
if, for every $i$ ($1\leq i\leq n$), $Z \models S_i\leq U_i$;
\item
the empty set, otherwise.
\end{itemize}
The {\sl reduct} $\Omega^Z$ of a program $\Omega$ with
respect to $Z$ is the union of the reducts of the rules of $\Omega$.

Consider, for example, the one-rule program
\begin{equation}
\label{ex1brule}
1 \leq \{a=2\}\leq 2 \ar 1\leq\{\no\ a=3,\no\ b=2\}\leq 4.
\end{equation}
Since the only head element of~(\ref{ex1brule}) is $a$, the reduct of
this rule
with respect to a set $Z$ of atoms is empty if $a \not\in Z$.  Consider the
case when $a \in Z$.  Since
$$Z\models\{\no\ a=3,\no\ b=2\}\leq 4,$$
the reduct consists of one rule
$$
a\ar (1\leq\{\no\ a=3,\no\ b=2\})^Z.
$$

It is clear from the definition of the reduct of a program above that every
rule in a reduct satisfies two conditions:
\begin{itemize}
 \item its head is a literal, and
 \item every member of its body has the form $L\leq S$ where $S$ does not
contain negative rule elements.
\end{itemize}
A rule satisfying these conditions is called a {\sl Horn rule}.
If a program $\Omega$ consists of Horn rules then there is a unique
minimal set $Z$ of literals such that $Z \models \Omega$. This set is called
the {\sl deductive closure} of $\Omega$ and denoted by $\ii{cl}(\Omega)$.

Finally, a consistent set $Z$ of literals is an {\sl answer set} for a
program
$\Omega$ if $Z \models \Omega$ and $\ii{cl}(\Omega^Z)=Z$.

To illustrate this definition, assume that~$\Omega$ is~(\ref{ex1arule}).
Set $\{a,b\}$ is not an answer
set for $\Omega$ because it does not satisfy $\Omega$.  Let us check that
every proper subset of $\{a,b\}$ is an answer set.  Clearly, every such
subset
satisfies $\Omega$.  It remains to show that each of these sets is the
deductive closure of the corresponding reduct of $\Omega$.
\begin{itemize}
\item
$\Omega^\emptyset$ is empty, so that $\ii{cl}(\Omega^\emptyset)=\emptyset$.
\item
$\Omega^{\{a\}}$ consists of the single rule $a$, so that
$\ii{cl}(\Omega^{\{a\}})={\{a\}}$.
\item
$\Omega^{\{b\}}$ consists of the single rule $b$, so that
$\ii{cl}(\Omega^{\{b\}})={\{b\}}$.
\end{itemize}

To give another example, let $\Omega$ be~(\ref{ex1brule}).  Set $\{b\}$ is
not an answer set for $\Omega$ because it does not satisfy $\Omega$.  The
other subsets of $\{a,b\}$ satisfy $\Omega$. Consider the corresponding
reducts.
\begin{itemize}
\item
$\Omega^\emptyset$ is empty, so that $\ii{cl}(\Omega^\emptyset)=\emptyset$.
\item
$\Omega^{\{a\}}$ is
$$
a\ar -1\leq \{\}.
$$
Consequently, $\ii{cl}(\Omega^{\{a\}})={\{a\}}$.
\item
$\Omega^{\{a,b\}}$ is
$$
a\ar 1\leq \{\}
$$
Consequently, $\ii{cl}(\Omega^{\{a,b\}})=\emptyset\neq\{a,b\}$.
\end{itemize}
We conclude that the answer sets for~(\ref{ex1brule}) are $\emptyset$
and $\{a\}$.

\section{Translations}\label{sec:translation}

\subsection{Basic Translation}\label{ssec:translation}

In this section, we give the main definition of this paper --- the
description
of a translation from the language of weight constraints to the language
of
nested expressions --- and state a theorem about the soundness of this
translation.  The definition of the translation consists of 4 parts.


\medskip\noindent{\sl 1. The translation of a constraint of the form
\begin{equation}
\label{weightlower}
L\leq \{ c_1 = w_1, \ldots, c_m = w_m\}
\end{equation}
is the nested expression
\begin{equation}
\label{lowerold}
\begin{array}c
\langle c_1,\dots,c_m\rangle :\left\{I: L\leq \sum_{i \in I} w_i\right\}
\end{array}
\end{equation}
where $I$ ranges over the subsets of $\{1, \ldots, m\}$.
We denote the translation of $L\leq S$ by $[L\leq S]$.}

\medskip\noindent{\sl 2.  The translation of a constraint of the form
\begin{equation}
\label{weightupper}
\{ c_1 = w_1, \ldots, c_m = w_m\} \leq U
\end{equation}
is the nested expression
\begin{equation}
\label{upperold}
\begin{array}c
\no\ \left(
   \langle c_1,\dots,c_m\rangle :\left\{I: U< \sum_{i \in I} w_i\right\}
\right).
\end{array}
\end{equation}
where $I$ ranges over the subsets of $\{1, \ldots, m\}$.
We denote the translation of \hbox{$S\leq U$} by $[S\leq U]$.}

\medskip\noindent{\sl 3. The translation of a general weight constraint
is defined by}
$$[L\leq S\leq U]=[L\leq S],[S\leq U].$$

Recall that $L \leq S$ is shorthand for $L \leq S\leq\infty$,
and $S \leq U$ is shorthand for $-\infty \leq S\leq U$; translations of
weight constraints of these special types have been defined earlier.
It is easy to see that the old definition of $[L\leq S]$ gives a nested
expression equivalent to $[L\leq S\leq\infty]$, and similarly for
$[S\leq U]$.

\medskip\noindent{\sl 4.  For any program~$\Omega$ with weight
constraints,
its translation $[\Omega]$ is the program with
nested expressions obtained from $\Omega$ by replacing
each rule~(\ref{rule}) with
\begin{equation}
\label{rulep}
(l_1; \no\ l_1),\ldots,(l_p; \no\ l_p),[C_0] \ar [C_1],\ldots,[C_n]
\end{equation}
where $l_1,\ldots,l_p$ are the positive head elements of~(\ref{rule}).}

\medskip
The conjunctive terms in $(l_1; \no\ l_1),\ldots,(l_p; \no\ l_p)$
express, intuitively, that we are free to decide about every positive
head element of the rule whether or not to include it in the answer set.


To illustrate this definition, let us apply it first to
program~(\ref{ex1arule}).  The translation of the cardinality
constraint~$0\leq \{a,b\}\leq 1$ is
\begin{equation}
[0\leq \{a,b\}],[\{a,b\}\leq 1].
\label{example3}
\end{equation}
The first conjunctive term is
$$\langle a,b \rangle : \{\emptyset,\{1\},\{2\},\{1,2\}\}$$
which equals
$$\top;a;b;(a,b)$$
and is equivalent to $\top$.  Similarly, the second conjunctive term is
equivalent to $\no\ (a,b)$.  Consequently,~(\ref{example3}) can be written
as $\no\ (a,b)$.  It follows that the translation of
program~(\ref{ex1arule})
can be written as
\begin{equation}
(a;\no\ a),(b ;\no\ b),\no\ (a , b).
\label{ex1atrans}
\end{equation}

Similarly, we can check that program~(\ref{ex1brule}) turns into
$$
a\ar (\no\ a;\no\ b),\no\ (\no\ a,\no\ b).
$$

The translation defined above is sound:

\begin{theorem}
\label{th1}
For any program $\Omega$ with weight constraints, $\Omega$ and
$[\Omega]$ have the same answer sets.
\end{theorem}


We will conclude this section with a few comments about translating
weight constraints of the forms $L\leq S$ and $S\leq U$.

In Section~\ref{sec:weight} we have agreed to identify
any rule element $c$ with the cardinality constraint~$1\leq \{c\}$,
and to drop the head of a rule with weight constraints when this head
is~$1\leq \{\ \}$.  It is easy to check that $[1\leq \{c\}]$
is equivalent to $c$, and~$[1\leq \{\ \}]$ is equivalent to $\bot$.

If the weights $w_1,\dots,w_m$ are integers then the inequality
in~(\ref{upperold}) is equivalent to
$\lfloor U\rfloor+1\leq \sum_{i \in I} w_i$.
Consequently, in the case of integer weights (in particular, in the case of
cardinality constraints), $[S\leq U]$ can be written
as~$\no\ [\lfloor U\rfloor+1\leq S]$.  This is similar to a
transformation that is used by the preprocessor {\sc lparse} of system
{\sc smodels}.

The sign $<$ in place of $\leq$ is not allowed in weight constraints.  But
sometimes it is convenient to write expressions of the form
$$
[L< \{ c_1 = w_1, \ldots, c_m = w_m\}]
$$
understood as shorthand for
\begin{equation}
\label{less}
\begin{array}c
\langle c_1,\dots,c_m\rangle :\left\{I: L<\sum_{i \in I} w_i\right\}.
\end{array}
\end{equation}
Using this notation, we can write $[S\leq U]$ as $\no\ [U<S]$.

Finally, note that each of the sets $X$ used in the
expressions $\langle c_1, \ldots, c_m  \rangle : X$
in formulas (\ref{lowerold}), (\ref{upperold}) and~(\ref{less}) 
satisfies the
assumption of Proposition~\ref{prop1} (Section~\ref{sec:abbr}), because the
weights $w_i$ are nonnegative.

\subsection{Nondisjunctive Translation}
                    \label{sec:nondisjunctive}

A rule with nested expressions (Section~\ref{sec:nested})
is {\sl nondisjunctive} if its head
is a literal or $\bot$. A {\sl nondisjunctive program} is a program with
nested expressions whose rules are nondisjunctive.

For any program~$\Omega$ with weight constraints, its
{\sl nondisjunctive translation} $[\Omega]^{nd}$ is the nondisjunctive
program obtained from $\Omega$ by replacing
each rule~(\ref{rule}) with $p+1$ rules
\begin{equation}
\label{nd}
\begin{array}l
l_j\ar \no\ \no\ l_j, [C_1],\ldots,[C_n] \qquad (1\leq j\leq p),\\
\bot \ar \no\ [C_0], [C_1],\ldots,[C_n],
\end{array}
\end{equation}
where $l_1,\ldots,l_p$ are the positive head elements of~(\ref{rule}).

For example, if $\Pi$ is~(\ref{ex1arule}) then $[\Pi]$, as we have seen,
is~(\ref{ex1atrans}); the nondisjunctive translation $[\Pi]^{nd}$ of the
same program is
\begin{equation}
\label{ex1and}
\begin{array}l
a\ar\no\ \no\ a,\\
b\ar\no\ \no\ b,\\
\bot\ar\no\ \no\ (a ; b).
\end{array}
\end{equation}

\begin{proposition}
\label{th2new}
For any program $\Omega$ with weight constraints, $[\Omega]^{nd}$ is
strongly equivalent to $[\Omega]$.
\end{proposition}

In combination with Theorem~\ref{th1}, this fact shows that the 
nondisjunctive
translation is sound: $\Omega$ and $[\Omega]^{nd}$ have the same answer 
sets.

Its proof is based on the following well-known fact
about intuitionistic logic:

\begin{fact}
\label{fact-i}
If $F$ is a propositional combination of formulas $F_1,\dots, F_m$
then $F\vee\neg F$ is intuitionistically derivable from
$F_1\vee\neg F_1$,$\dots$,$F_m\vee\neg F_m$.
\end{fact}

\noindent
\begin{proof}[Proof of Proposition~\ref{th2new}]
We will show that formula~(\ref{rulep}) is equivalent to the conjunction
of the formulas~(\ref{nd}) in the logic of here-and-there.  By
Fact~\ref{fact-i}, the formula
\begin{equation}
[C_0] \vee \neg [C_0]
\label{c0}
\end{equation}
is entailed by the formulas $c\vee\neg c$ for all head elements $c$ of
rule~(\ref{rule}).  For every negative~$c$, $c\vee\neg c$ is provable
in the logic of here-and-there.  It follows that (\ref{c0}) is derivable in
the logic of here-and-there from $l_1\vee\neg l_1$,$\dots$,$l_p\vee\neg l_p$.
Consequently, $\neg\neg[C_0] \equiv [C_0]$ is derivable from these formulas
as well.  Hence~(\ref{rulep}) is
equivalent in the logic of here-and-there to the rule
$$(l_1; \no\ l_1),\ldots,(l_p; \no\ l_p),\no\ \no\ [C_0]
                    \ar [C_1],\ldots,[C_n]$$
which can be broken into the rules
$$
\begin{array}l
l_j; \no\ l_j \ar [C_1],\ldots,[C_n] \qquad (1\leq j\leq p),\\
\no\ \no\ [C_0] \ar [C_1],\ldots,[C_n].
\end{array}
$$
The first line is equivalent to the first line of~(\ref{nd})
in the logic of here-and-there.
The second line is intuitionistically equivalent to the second line
of~(\ref{nd}).
\end{proof}

\subsection{Eliminating Nested Expressions}
                    \label{sec:nonnested}

A nondisjunctive rule is {\sl nonnested} if its body is a conjunction of
literals, each possibly prefixed with $\no$.  A {\sl nonnested program} is
a program whose rules are nonnested.  Thus the syntactic form of nonnested
programs is the same as in the simple case
reviewed at the beginning of Introduction, except that the head of a
nonnested rule can be $\bot$, and that literals are allowed in place of 
atoms.

Since the answer sets for a nonnested program have the anti-chain property,
turning a program with
weight constraints into a nonnested program with the same answer sets is,
generally, impossible.  But we can turn any program with weight constraints
into its nonnested conservative extension---into a program
that may contain new atoms; dropping the new atoms from the answer sets
of the translation gives the answer sets for the original program.

Each of the new atoms introduced in the nonnested translation
$[\Omega]^{nn}$ below is, intuitively, an ``abbreviation'' for some formula
related to the nondisjunctive translation $[\Omega]^{nd}$.
For instance, to eliminate the nesting of negations from the first line of
the nondisjunctive translation~(\ref{nd}), we will introduce, for every $j$,
a new atom $q_{\sno\ l_j}$, and replace that line with the rules
$$
\begin{array}l
q_{\sno\ l_j}\ar\no\ l_j,\\
l_j\ar \no\ q_{\sno\ l_j}, [C_1],\ldots,[C_n]
\end{array}
$$
($1\leq j\leq p$).  The first of these rules tells us that the new atom
$q_{\sno\ l_j}$ is used to ``abbreviate'' the formula $\no\ l_j$.  The
second rule is the first of rules~(\ref{nd}) with this subformula 
replaced by
the corresponding atom.  For instance, the nondisjunctive
translation~(\ref{ex1and}) of program~(\ref{ex1arule}) turns after this
transformation into
\begin{equation}
\label{ex1andqnot}
\begin{array}l
q_{\sno\ a}\ar\no\ a,\\
a\ar\no q_{\sno\ a}, \\
q_{\sno\ b}\ar\no\ b,\\
b\ar\no\ q_{\sno\ b}, \\
\bot\ar\no\ \no\ (a ; b).
\end{array}
\end{equation}

Introducing the atoms $q_{\sno\ l_j}$ brings us very close to the goal
of eliminating  nesting altogether, because every rule of the program
obtained from $[\Omega]^{nd}$ by this transformation is strongly equivalent
to a set of nonnested rules.  One way to eliminate nesting is to convert the
body of every rule to a ``disjunctive normal form'' using
De Morgan laws, the distributivity of conjunction over disjunction, and, in
case of the second line of~(\ref{nd}), double negation
elimination.\footnote{Double negation elimination in the body  of a rule
with the head $\bot$ is intuitionistically valid.} After
that, we can break every rule into several nonnnested rules, each
corresponding to one of the disjunctive terms of the body.  For 
instance, the
last rule of~(\ref{ex1andqnot}) becomes
$$
\bot\ar a ; b
$$
after the first step and
$$
\begin{array}l
\bot\ar a,\\
\bot\ar b
\end{array}
$$
after the second.

The definition of $[\Omega]^{nn}$ below follows a different approach to the
elimination of the remaining nested expressions.  Besides the ``negation
atoms'' of the form $q_{\sno\ l_j}$, it introduces other new atoms, to make the
translation of weight constraints more compact in some cases.
These ``weight atoms'' have the forms
$q_{w\leq S}$ and $q_{w< S}$, where $w$ is a number and $S$ is an
expression of the form $\{c_1=w_1,\dots,c_m=w_m\}$ for some rule
elements $c_1,\dots,c_m$ and nonnegative numbers $w_1,\dots,w_m$.
They ``abbreviate'' the formulas $[w\leq S]$ and $[w<S]$ respectively.

In the following definition, $\{c_1=w_1,\dots,c_m=w_m\}'$, where $m>0$, stands
for $\{c_1=w_1,\dots,c_{m-1}=w_{m-1}\}$.
Consider a nonnested program $\Pi$ that may contain atoms of the forms
$q_{w\leq S}$ and $q_{w< S}$.  We say that
$\Pi$ is {\sl closed} if
\begin{itemize}
\item for each atom of the form $q_{w\leq S}$ that occurs in $\Pi$,
$\Pi$ contains the rule
\begin{equation}
\label{lower1}
         q_{w\leq S}
\end{equation}
if $w\leq 0$, and the pair of rules
\begin{equation}
\label{middle1}
\begin{array}l
         q_{w\leq S} \leftarrow q_{w\leq S'},\\
         q_{w\leq S} \leftarrow c_m,q_{w-w_m\leq S'}
\end{array}
\end{equation}
if $0<w\leq w_1+\dots+w_m$;
\item for each atom of the form $q_{w<S}$ that occurs in $\Pi$,
$\Pi$ contains the rule
\begin{equation}
\label{lower2}
         q_{w< S}
\end{equation}
if $w<0$, and the pair of rules
\begin{equation}
\label{middle2}
\begin{array}l
         q_{w< S} \leftarrow q_{w< S'},\\
         q_{w< S} \leftarrow c_m,q_{w-w_m < S'}
\end{array}
\end{equation}
if $0\leq w < w_1+\dots+w_m$.
\end{itemize}

We define the nonnested translation $[L\leq S\leq U]^{nn}$ of a weight
constraint $L\leq S\leq U$ as the conjunction
$$q_{L\leq S}, \no\ q_{U<S}.$$

Now we are ready to define the nonnested translation of a program.  For any
program $\Omega$ with weight constraints, $[\Omega]^{nn}$  is the smallest
closed program that contains, for every rule
$$
      L_0\leq S_0\leq U_0 \leftarrow C_1,\dots,C_n
$$
of $\Omega$, the rules
\begin{equation}
\label{choice-a}
      q_{\sno\ l} \ar \no\ l
\end{equation}
and
\begin{equation}
\label{choice-b}
      l \ar \no\ q_{\sno\ l},[C_1]^{nn},\dots,[C_n]^{nn}
\end{equation}
for each of its positive head elements $l$, and the rules
\begin{equation}
\label{constraint}
\begin{array}l
          \bot \ar \no\ q_{L_0\leq S_0},[C_1]^{nn},\dots,[C_n]^{nn},\\
          \bot \ar q_{U_0<S_0},[C_1]^{nn},\dots,[C_n]^{nn}.
\end{array}
\end{equation}

For instance, if $\Omega$ is~(\ref{ex1arule}) then
rules~(\ref{choice-a})--(\ref{constraint}) are
\begin{equation}
\label{ex1ann}
\begin{array}l
q_{\sno\ a}\ar\no\ a,\\
a\ar\no\ q_{\sno\ a}, \\
q_{\sno\ b}\ar\no\ b,\\
b\ar\no\ q_{\sno\ b}, \\
\bot\ar\no\ q_{0\leq\{a,b\}},\\
\bot\ar q_{1<\{a,b\}}.
\end{array}
\end{equation}
To make this program closed, we add to it the following ``definitions''
of the weight atoms $q_{0\leq\{a,b\}}$ and $q_{1<\{a,b\}}$, and, recursively,
of the weight atoms that are used in these definitions:
\begin{equation}
\label{ex1anndefs}
\begin{array}l
q_{0\leq \{a,b\}},\\
q_{1< \{a,b\}} \ar q_{1< \{a\}},\\
q_{1< \{a,b\}} \ar b,q_{0< \{a\}},\\
q_{0< \{a\}}   \ar q_{0< \{\}},\\
q_{0< \{a\}}   \ar a,q_{-1< \{\}},\\
q_{-1< \{\}}.\\
\end{array}
\end{equation}
The nonnested translation of~(\ref{ex1arule}) consists of
rules~(\ref{ex1ann}) and~(\ref{ex1anndefs}).

The following theorem describes the relationship between the answer sets for
$\Omega$ and the answer sets for $[\Omega]^{nn}$.  In the statement of the
theorem, $Q_\Omega$ stands for the set of all new atoms that occur
in~$[\Omega]^{nn}$---both negation atoms $q_{\sno\ l}$ and weight atoms
$q_{w\leq S}$, $q_{w<S}$.

\begin{theorem}
\label{th2}
For any program $\Omega$ with weight constraints,
$Z\mapsto Z \setminus Q_\Omega$ is a 1--1 correspondence between the
answer sets for $[\Omega]^{nn}$ and the answer sets for $\Omega$.
\end{theorem}

Recall that the introduction of the new atoms $q_{w\leq S}$ and $q_{w< S}$
is motivated by the desire to make the translations of programs
more compact.  We will investigate now
to what degree this goal has been achieved.

The basic translation $[C]$ of a weight constraint, as defined in
Section~\ref{ssec:translation}, can be exponentially
larger than $C$.  For this reason, the basic and nondisjunctive
translations of a program~$\Omega$ are, generally, exponentially larger
than~$\Omega$.

The nonnested translation of a program $\Omega$ consists of
the rules~(\ref{choice-a})--(\ref{constraint}) corresponding to all rules
of~$\Omega$, and the additional rules~(\ref{lower1})--(\ref{middle2})
that make the program closed.
The part consisting of rules~(\ref{choice-a})--(\ref{constraint})
cannot be significantly larger than~$\Omega$, because each
of the formulas~$[C_i]^{nn}$ is short --- it contains at most two atoms.
The second part consists of the ``definitions'' of all weight atoms in
$[\Omega]^{nn}$, and it contains at most two short rules for every such atom.
Under what conditions can we guarantee that the number of weight atoms is not
large in comparison with the size of~$\Omega$?

The {\sl length} of a weight constraint~(\ref{wc}) is~$m$, and its 
{\sl weight} is $w_1+\cdots+w_m$.  We will denote the length of $C$ by $L(C)$,
and the weight of $C$ by $W(C)$.

\begin{proposition}
\label{prop2}
For programs $\Omega$ without non-integer weights, the number of weight
atoms occurring in $[\Omega]^{nn}$ is $O\big(\sum L(C)\cdot W(C)\big)$,
where the sum extends over all weight constraints $C$ occurring
in~$\Omega$.
\end{proposition}

If the weights in $\Omega$ come from a fixed finite set of integers
(for instance, if every weight constraint in $\Omega$ is a cardinality
constraint) then $W(C)=O(L(C))$, and the proposition above shows that
the number of weight atoms in $[\Omega]^{nn}$ is not large in comparison
with the size of $\Omega$.  Consequently, in this case $[\Omega]^{nn}$ cannot
be large in comparison with $\Omega$ either.

\noindent{\samepage
\begin{proof}[\sl Proof of Proposition~\ref{prop2}]
Let~$\Omega$ be a program without non-integer weights.  About a rule
from~$[\Omega]^{nn}$ we will say that it is {\sl relevant} if for every
weight atom $w\leq S$ or $w<S$ occurring in that rule there is a weight
constraint~(\ref{wc}) in~$\Omega$ such that $S$ is
$\{ c_1 = w_1, \ldots, c_j = w_j\}$ for some $j\in\{0,\dots,m\}$, and
$$w\in\{-\max(w_1,\dots,w_m),\dots,w_1+\dots+w_m\}\cup\{L,U\}.$$
It is clear that the number of weight atoms occurring in relevant rules can
be estimated as $O\big(\sum L(C)\cdot W(C)\big)$.  On the other hand, it is
easy to see that the set of relevant rules contains the
rules~(\ref{choice-a})--(\ref{constraint})  corresponding to all rules
of~$\Omega$, and that it is closed.  Consequently, all rules in
$[\Omega]^{nn}$ are relevant.
\end{proof}
}

Without the assumption that all weights in $\Omega$ are integers, we
can guarantee that the number of weight atoms occurring in
$[\Omega]^{nn}$ is $O\big(\sum 2^{L(C)}\big)$.

\section{Proving Strong Equivalence of Programs with Weight Constraints}
\label{sec:applications}

For programs with weight constraints, the definition of strong equivalence
is similar to the definition given above:  $\Omega_1$ and $\Omega_2$ are
{\sl strongly equivalent} to each other if, for every program $\Omega$ with
weight constraints, the union $\Omega_1\cup \Omega$ has the
same answer sets as  $\Omega_2\cup \Omega$.  The method of proving strong
equivalence of programs with weight constraints discussed in this section
is based on the following proposition:

\begin{proposition}
\label{prop-strong}
$\Omega_1$ is strongly equivalent to $\Omega_2$ iff~$[\Omega_1]$ is
strongly equivalent to~$[\Omega_2]$.
\end{proposition}

\noindent\begin{proof}
Assume that $[\Omega_1]$ is strongly equivalent to $[\Omega_2]$.  Then, for
any program with weight constraints $\Omega$, $[\Omega_1]\cup [\Omega]$ has the
same answer sets as  $[\Omega_2]\cup [\Omega]$.  The first program equals
$[\Omega_1\cup\Omega]$, and, by Theorem~\ref{th1}, has the same answer sets
as $\Omega_1\cup\Omega$.  Similarly, the second program has the same answer
sets as $\Omega_2\cup\Omega$.  Consequently $\Omega_1$ is strongly
equivalent to $\Omega_2$.

Assume now that $[\Omega_1]$ is not strongly equivalent to $[\Omega_2]$.
Consider the corresponding programs $[\Omega_1]'$, $[\Omega_2]'$ without
classical negation, formed as described at the end of Section~\ref{sec:ht},
and let \ii{Cons}
be the set of formulas $\neg(a\wedge a')$ for all new atoms $a'$ occurring
in these programs.  By Theorem~2 from \cite{lif01}, $[\Omega_1]'\cup\ii{Cons}$
is not equivalent to $[\Omega_2]'\cup\ii{Cons}$ in the logic of here-and-there.
It follows by Theorem 1 from~\cite{lif01} that there exists a unary
program~$\Pi$ such that $[\Omega_1]'\cup\ii{Cons}\cup\Pi$ and
$[\Omega_2]'\cup\ii{Cons}\cup\Pi$ have different collections of answer sets.
(A program with nested expressions is said to be unary if each of its rules
is an atom or has the form $a_1\ar a_2$ where $a_1$, $a_2$ are atoms.)
Let $\Pi^*$ be the program obtained from $\Pi$ by replacing each atom of the
form $a'$ by $\neg a$.  In view of the convention about identifying any
literal~$l$ with the weight constraint $1\leq\{l=1\}$
(Section~\ref{sec:syntaxweight}), $\Pi^*$
can be viewed as a program with weight constraints, and it's easy to check that
$[\Pi^*]'$ is strongly equivalent to $\Pi$.  Then, for $i=1,2$, the program
$[\Omega_i]'\cup\ii{Cons}\cup\Pi$  has the same answer sets as the program
$[\Omega_i]'\cup\ii{Cons}\cup[\Pi^*]'$, which can be rewritten as
$[\Omega_i\cup\Pi^*]'\cup\ii{Cons}$.  By the choice of $\Pi$,
it follows that the collection of answer sets of
$[\Omega_1\cup\Pi^*]'\cup\ii{Cons}$ is different from the collection of
answer sets of
$[\Omega_2\cup\Pi^*]'\cup\ii{Cons}$.  Consequently, the same can be said
about the pair of programs $[\Omega_1\cup\Pi^*]$ and $[\Omega_2\cup\Pi^*]$,
and, by Theorem~\ref{th1}, about $\Omega_1\cup\Pi^*$ and $\Omega_2\cup\Pi^*$.
It follows that $\Omega_1$ is not strongly equivalent to $\Omega_2$.
\end{proof}

As an example, let us check that the program
\begin{equation}
\label{exseq1}
\begin{array}l
1\leq\{p,q\}\leq 1\\
p
\end{array}
\end{equation}
is strongly equivalent to
\begin{equation}
\label{exseq2}
\begin{array}l
\ar q\\
p.
\end{array}
\end{equation}
Rules~(\ref{exseq1}), translated into the language of nested expressions
and written in the syntax of propositional formulas, become
$$
\begin{array}l
(p\vee\neg p)\wedge(q\vee\neg q)\wedge(p\vee q)\wedge\neg(p\wedge q)\\
(p\vee\neg p)\wedge p.
\end{array}
$$
Rules~(\ref{exseq2}), rewritten in a similar way, become
$$
\begin{array}l
\neg q\\
(p\vee\neg p)\wedge p.
\end{array}
$$
It is clear that each of these sets of formulas is intuitionistically
equivalent to $\{p,\neg q\}$.

The fact that programs~(\ref{exseq1}) and~(\ref{exseq2}) are strongly
equivalent to each other can be also proved directly, using the definition
of strong equivalence and the definition of an answer set for programs with
weight constraints.  But this proof would not be as easy as the one above.
Generally, to establish that a program $\Omega_1$ is strongly equivalent to
a program $\Omega_2$, we need to show that for every program $\Omega$ and
every consistent set~$Z$ of literals,
\begin{enumerate}
\item[(a$_1$)] $Z \models \Omega_1\cup\Omega$ and
\item[(b$_1$)] $\ii{cl}((\Omega_1\cup\Omega)^Z)=Z$
\end{enumerate}
if and only if
\begin{enumerate}
\item[(a$_2$)] $Z \models \Omega_2\cup\Omega$ and
\item[(b$_2$)] $\ii{cl}((\Omega_2\cup\Omega)^Z)=Z$.
\end{enumerate}
Sometimes we may be able to check separately that
(a$_1$) is equivalent to (a$_2$) and that (b$_1$) is equivalent to (b$_2$),
but in other cases this may not work.  For instance,
if~$\Omega_1$ is~(\ref{exseq1}) and~$\Omega_2$ is~(\ref{exseq2}) then
(b$_1$) may not be equivalent to (b$_2$).

An alternative method of establishing the strong equivalence of programs
with weight constraints is proposed in~\cite[Section~6]{tur03}.  According to
that approach, we check that for every consistent set $Z$ of literals
and every subset $Z'$ of $Z$,
\begin{enumerate}
\item[(a$_3$)] $Z \models \Omega_1$ and
\item[(b$_3$)] $Z' \models \Omega_1^Z$
\end{enumerate}
if and only if
\begin{enumerate}
\item[(a$_4$)] $Z \models \Omega_2$ and
\item[(b$_4$)] $Z' \models \Omega_2^Z$.
\end{enumerate}

\section{Proofs of Theorems}\label{sec:proof}

\subsection{Proof of Theorem~\ref{th1}}

\begin{lemma}
\label{newlemma}
For any weight constraint $C$ and any consistent set $Z$ of literals,
\hbox{$Z\models [C]$} iff $Z\models C$.
\end{lemma}

\noindent
\begin{proof*}
It is sufficient to prove the assertion of the lemma for constraints of the
forms $L\leq S$ and $S\leq U$.  Let $S$ be $\{ c_1 = w_1, \ldots, c_m =
w_m\}$.
Then, by Proposition~\ref{prop1} (Section~\ref{sec:abbr}),
$$
\begin{array}{rcl}
Z\models [L\leq S] &\hbox{iff}&{
        \{i: Z\models c_i\}\in \left\{I: L\leq \sum_{i \in I}
w_i\right\}}\\ \\
     &\hbox{iff}&{
        L\leq \sum_{i: Z\models c_i} w_i}\\ \\
     &\hbox{iff}& Z\models L\leq S.
\end{array}
$$
Similarly,
$$
\begin{array}{rcl}
Z\models [S\leq U] &\hbox{iff}&{
        \{i: Z\models c_i\}\not \in \left\{I: U< \sum_{i \in I}
w_i\right\}}\\ \\
     &\hbox{iff}&{
        U\geq \sum_{i: Z\models c_i} w_i}\\ \\
     &\hbox{iff}& Z\models S\leq U. \mathproofbox
\end{array}
$$
\end{proof*}

\begin{lemma}
\label{proplower}
For any constraint $L\leq S$ and any consistent sets $Z$, $Z'$ of literals,
$$
Z'\models [L\leq S]^Z \text{ iff } Z'\models (L\leq S)^Z.
$$
\end{lemma}

\noindent\begin{proof}
Let $S$ be $\{ c_1 = w_1, \ldots, c_m = w_m\}$ and let $I$ stand for
$\{1,\dots,m\}$.  It is immediate from the definition of the reduct in
Section~\ref{sec:semanticsnested} that
\begin{equation}
\label{newred}
\big(\langle F_1,\dots,F_n\rangle :X\big)^Z =
        \langle F_1^Z,\dots,F_n^Z\rangle:X.
\end{equation}
For any subset $J$ of $I$, let $\Sigma J$ stand for $\sum_{i\in J} w_i$.
Using~(\ref{newred}) and Proposition~\ref{prop1}, we can rewrite the
left-hand side of the equivalence to be proved as follows:
$$
\begin{array}{rcl}
Z'\models [L\leq S]^Z &\hbox{iff} &{
    Z'\models \langle c_1^Z,\dots,c_m^Z\rangle :
              \left\{J\subseteq I: L\leq \Sigma J\right\}}\\ \\
              &\hbox{iff}&{
    \{i\in I:Z'\models c_i^Z\}\in
        \left\{J\subseteq I: L\leq \Sigma J\right\}}\\ \\
              &\hbox{iff}&{
    L\leq \Sigma\{i\in I:Z'\models c_i^Z\}}
\end{array}
$$
Let $I'$ be the set of all $i\in I$ such that the rule element~$c_i$ is
positive, and let~$I''$ be the set of all $i\in I\setminus I'$ such that
$Z\models c_i$.
It is clear that $c_i^Z$ is $c_i$ for $i\in I'$, $\top$ for $i\in I''$,
and $\bot$ for all other values of $i$.  Consequently
$$
\begin{array}{rcl}
Z'\models [L\leq S]^Z &\hbox{iff} &{
    L\leq \Sigma\{i\in I':Z'\models c_i\}+\Sigma I''}\\ \\
               &\hbox{iff}&{
    L-\Sigma I''\leq \Sigma\{i\in I':Z'\models c_i\}}\\ \\
               &\hbox{iff}&{
    Z'\models (L^Z\leq S')}
\end{array}$$
where $L^Z$ and $S'$ are defined as in Section~\ref{sec:semanticsweight}.
It remains to notice that
$(L\leq S)^Z=(L^Z\leq S')$.
\end{proof}

\begin{lemma}
\label{propupper}
For any constraint $S\leq U$ and any consistent set $Z$ of literals,
$$
[S\leq U]^Z =
   \begin{cases}
\top\ , & \text{if $Z\models (S \leq U)$}, \cr
\bot\ , & \text{otherwise}. \hfill \end{cases}
$$
\end{lemma}

\noindent\begin{proof}
By the definition of the reduct in Section~\ref{sec:semanticsnested},
$[S\leq U]^Z$ is
\begin{enumerate}[$\bullet$]
 \item[$\bullet$]$\top$, if $Z \not \models [U< S]$,
 \item[$\bullet$] $\bot$, otherwise.
\end{enumerate}
It remains to notice that $Z \not \models [U< S]$ iff 
$Z\models [S\leq U]$, and then iff $Z\models S\leq U$
by Lemma~\ref{newlemma}.
\end{proof}

In Lemmas~\ref{answerseteq}--\ref{closureequiv}, $\Omega$ is an arbitrary
program with weight constraints.  Recall that, according to
Section~\ref{sec:nondisjunctive}, the nondisjunctive
translation~$[\Omega]^{nd}$ of~$\Omega$ consists of rules of two kinds:
\begin{equation}
\label{rulep1a}
l_j\ar \no\ \no\ l_j, [C_1],\ldots,[C_n]
\end{equation}
and
\begin{equation}
\label{rulep1c}
\bot \ar \no\ [C_0], [C_1],\ldots,[C_n].
\end{equation}
We will denote the set of rules~(\ref{rulep1a}) corresponding to all 
rules of
$\Omega$ by $\Pi_1$,
and the set of rules~(\ref{rulep1c}) corresponding to all rules of $\Omega$
by $\Pi_2$, so that
\begin{equation}
\label{union}
[\Omega]^{nd} = \Pi_1\cup\Pi_2.
\end{equation}

\begin{lemma}
\label{answerseteq}
A consistent set $Z$ of literals is an answer set for $[\Omega]^{nd}$ 
iff $Z$
is an answer set for $\Pi_1$ and $Z\models \Pi_2$.
\end{lemma}

In view of~(\ref{union}), this is an instance of a general fact,
proved in~\cite{lif99d} as Proposition~2, that can be restated as the
following:
\begin{fact}
\label{fact0a}
Let $\Pi_1$, $\Pi_2$ be programs with nested expressions such that the
head of every rule in $\Pi_2$ is $\bot$.  A consistent set $Z$ of
literals is
an answer set for $\Pi_1\cup\Pi_2$ iff $Z$ is an answer set for $\Pi_1$ and
$Z\models \Pi_2$.
\end{fact}

\begin{lemma}  For any consistent set $Z$ of literals,
\label{satisf}
$Z\models \Omega$ iff $Z\models \Pi_2$.
\end{lemma}

\noindent\begin{proof}
It is sufficient to consider the case when $\Omega$ consists of a single
rule~(\ref{rule}).  In this case, $Z\models \Omega$ iff
$$Z\models C_0\hbox{ or, for some }i\ (1\leq i\leq m),\ Z\not\models C_i.$$
On the other hand, $Z\models \Pi_2$ iff
$$Z\models [C_0]\hbox{ or, for some }i\ (1\leq i\leq m),\ Z\not\models
[C_i].$$
By Lemma~\ref{newlemma}, these conditions are equivalent to each other.
\end{proof}

\begin{lemma}
For any consistent sets $Z$, $Z'$ of literals,
\label{translreduct}
$Z'\models \Omega^Z$ iff $Z'\models \Pi_1^Z$.
\end{lemma}

\noindent\begin{proof}
It is sufficient to consider the case when $\Omega$ consists of a single
rule~(\ref{weightrule}).  Then $\Pi_1^Z$ consists of the rules
\begin{equation}
\label{tr1}
l \ar (\no\ \no\ l)^Z,[L_1\leq S_1]^Z,[S_1\leq U_1]^Z,
                   \ldots,[L_n\leq S_n]^Z,[S_n\leq U_n]^Z
\end{equation}
for all positive head elements $l$ of~(\ref{weightrule}).

{\sl Case 1:} for every $i$ ($1\leq i\leq n$),
$Z \models S_i\leq U_i$.  Then, by Lemma~\ref{propupper}, each of the
formulas
$[S_1\leq U_1]^Z,\dots,[S_n\leq U_n]^Z$ is $\top$.  Note also that
if $l\not\in Z$ then $(\no\ \no\ l)^Z$ is $\bot$, so that~(\ref{tr1})
is satisfied by any consistent set of literals.  Consequently $Z'$
satisfies $\Pi_1^Z$ iff, for each positive head element $l\in Z$,
\begin{equation}
\label{z1}
Z'\models l\hbox{ or, for some }
i\ (1\leq i\leq m),\ Z'\not\models [L_i\leq S_i]^Z.
\end{equation}
On the other hand, according to the definition of the reduct
from Section~\ref{sec:semanticsweight}, $\Omega^Z$ is the set of rules
$$
l \ar (L_1\leq S_1)^Z,\ldots,(L_n\leq S_n)^Z
$$
for all positive head elements $l$ satisfied by $Z$.
Then $Z'\models \Omega^Z$ iff, for each positive head element $l\in Z$,
$$
Z'\models l\hbox{ or, for some }
i\ (1\leq i\leq m),\ Z'\not\models (L_i\leq S_i)^Z.
$$
By Lemma~\ref{proplower}, this condition is equivalent to~(\ref{z1}).

{\sl Case 2:} for some $i$, $Z \not\models S_i\leq U_i$.
Then, by Lemma~\ref{propupper}, one of the formulas
$[S_i\leq U_i]^Z$ is $\bot$, so
that each rule~(\ref{tr1}) is trivially satisfied by any $Z'$.
On the other hand, in this case $\Omega^Z$ is empty.
\end{proof}

\begin{lemma}
\label{closureequiv}
If set $\ii{cl}(\Omega^Z)$ is consistent then it is the only answer
set for~$\Pi_1^Z$; otherwise,~$\Pi_1^Z$ has no answer sets.
\end{lemma}

\noindent\begin{proof}
Recall that $\ii{cl}(\Omega^Z)$ is defined as the unique minimal set
satisfying~$\Omega^Z$ (Section~\ref{sec:semanticsweight}).  The answer sets for
a program with nested expressions that does not contain negation as failure
are defined as the minimal consistent sets satisfying that
program (Section~\ref{sec:semanticsnested}).  It remains to notice that
$\Omega^Z$ and $\Pi_1^Z$ are satisfied by the same sets of literals
(Lemma~\ref{translreduct}).
\end{proof}

\begin{theorem2}
For any program $\Omega$ with weight constraints, $\Omega$ and
$[\Omega]$ have the same answer sets.
\end{theorem2}

\noindent\begin{proof}
By the definition of an answer set for programs with weight constraints
(Section~\ref{sec:weight}), a consistent set $Z$ of literals is an answer
set for $\Omega$ iff
$$
\ii{cl}(\Omega^Z)=Z\hbox{ and }Z\models \Omega.
$$
By Lemmas~\ref{closureequiv} and~\ref{satisf}, this is equivalent to the
condition
$$
Z\hbox{ is an answer set for $\Pi_1^Z$ and }Z\models \Pi_2.
$$
By the definition of an answer set for programs with nested expressions
(Section~\ref{sec:nested}) and by Lemma~\ref{answerseteq}, this is further
equivalent to saying that $Z$ is an answer set for $[\Omega]^{nd}$.  By
Proposition~\ref{th2new}, $[\Omega]^{nd}$ has the same answer sets as
$[\Omega]$.
\end{proof}

\subsection{Two Lemmas on Programs with Nested Expressions}\label{sec:lemmas}

The idea of program completion~\cite{cla78} is that the set of rules of a
program with the same atom $q$ in the head is the ``if'' part of a definition
of~$q$; the ``only if'' half of that definition is left implicit.
If, for instance, the rule
$$q\leftarrow F$$
is the only rule in the program whose head is $q$ then that rule is an
abbreviated form of the assertion that $q$ is equivalent to $F$.

Since in a rule with nested expressions the head is allowed to have
the same syntactic structure as the body, the ``only if'' part of such an
equivalence can be expressed by a rule also:
$$F\leftarrow q.$$
The lemma below shows that adding such rules to a program does not change its
answer sets.

An occurrence of a formula $F$ in a formula or a rule is
{\sl singular} if the symbol before this occurrence of $F$ is $\neg$;
otherwise the occurrence is {\sl regular}~\cite{lif99d}.  The expression
$$
F\leftrightarrow G
$$
stands for the pair of rules
$$
\begin{array}l
F\leftarrow G\\
G\leftarrow F.
\end{array}
$$

  \intheoremtrue
  \normalfont\rmfamily
  \trivlist
    \pagebreak[3]\item[\hskip \labelsep{\normalfont\itshape Completion Lemma}]%
    \item[]%
Let $\Pi$ be a program with nested expressions, and let $Q$ be a set of
atoms that do not have regular occurrences in the heads of the rules of $\Pi$.
For every $q\in Q$, let $\ii{Def}(q)$ be a formula.  Then the program
$$\Pi\cup\{q\leftarrow \ii{Def}(q)\; : \;q\in Q\}$$
has the same answer sets as the program
$$\Pi\cup\{q\leftrightarrow \ii{Def}(q)\; : \;q\in Q\}.$$
  \endtrivlist\intheoremfalse

In the special case when $Q$ is a singleton this fact was first
proved by Esra Erdem (personal communication).

In the statement of the completion lemma, if the atoms from~$Q$ occur
neither in~$\Pi$ nor in the formulas $\ii{Def}(q)$ then adding the
rules $q\leftarrow \ii{Def}(q)$ to $\Pi$ extends the
program by ``explicit definitions'' of ``new'' atoms.  According to the lemma
below, such an extension is conservative: the answer sets for $\Pi$ can be
obtained by dropping the new atoms from the answer sets for the extended
program.

  \intheoremtrue
  \normalfont\rmfamily
  \trivlist
    \pagebreak[3]\item[\hskip \labelsep{\normalfont\itshape Lemma on Explicit Definitions}]%
    \item[]%
Let $\Pi$ be a program with nested expressions, and let $Q$ be a set of
atoms that do not occur in~$\Pi$.  For every $q\in Q$, let $\ii{Def}(q)$ be
a formula that contains no atoms from $Q$.  Then $Z\mapsto Z \setminus Q$
is a 1--1 correspondence between the answer sets for
$\Pi\cup\{q\leftarrow \ii{Def}(q)\; : \;q\in Q\}$
and the answer sets for~$\Pi$.
  \endtrivlist\intheoremfalse

The completion lemma and the lemma on explicit definitions can be proved
as follows.

\begin{lemma}
\label{monotone}
Let $\Pi$ be a program without negation as failure, and $Z'$ a subset of
a consistent set $Z$ of literals.  If the literals in $Z\setminus Z'$ do
not have
regular occurrences in the heads of the rules of $\Pi$ and $Z\models \Pi$
then $Z'\models \Pi$.
\end{lemma}

The proof of this lemma uses the following fact that is easy to verify
by structural induction:

\begin{fact}
\label{fact2}
Let $F$ be a formula without negation as failure, $Z$ a consistent set of
literals and $Z'$ a subset of $Z$. If $Z'\models F$ then $Z\models F$.
\end{fact}

\noindent\begin{proof}[Proof of Lemma~\ref{monotone}]
Take a rule $\ii{Head} \ar \ii{Body}$ in~$\Pi$  such that
$Z'\models \ii{Body}$.  By  Fact~\ref{fact2}, $Z\models \ii{Body}$, and
consequently $Z\models \ii{Head}$.  Since the literals in $Z\setminus Z'$
do not have regular occurrences in \ii{Head}, it follows that
$Z'\models \ii{Head}$.
\end{proof}

\begin{lemma}
\label{samereduct}
Let $\Pi$ be a logic program, and let $S$ be the set of literals
that have regular occurrences in $\Pi$ in the scope of negation as failure.
For any pair $Z_1$, $Z_2$ of consistent sets of literals, if
$Z_1\cap S=Z_2\cap S$ then $\Pi^{Z_1}=\Pi^{Z_2}$.
\end{lemma}

\noindent\begin{proof}
>From the condition $Z_1\cap S=Z_2\cap S$ we conclude that for every formula~$F$
occurring in $\Pi$ in the scope of negation as failure, $Z_1\models F$
iff $Z_2\models F$. Then the fact that $F^{Z_1}=F^{Z_2}$ for every formula~$F$
occurring in~$\Pi$ follows by structural induction.
\end{proof}

\noindent\begin{proof}[Proof of the Completion Lemma]
First consider the case when $\Pi$ and the formulas $\ii{Def}(q)$ do not
contain negation as failure; the general case is discussed at the end of
the proof.  Let $\Pi_1$ stand for
$\Pi\cup\{q\leftarrow \ii{Def}(q)\; : \;q\in Q\}$, and $\Pi_2$ stand for
$\Pi\cup\{q\leftrightarrow \ii{Def}(q)\; : \;q\in Q\}$.
We need to show that $Z$ is minimal among the sets satisfying $\Pi_1$
iff $Z$ is minimal among the sets satisfying $\Pi_2$.

\medskip\noindent{\sl Case 1:}
For every subset $Z'$ of $Z$, if $Z'\models \Pi_1$
then $Z'\models \Pi_2$.  The opposite holds also, because
$\Pi_1\subseteq \Pi_2$.  Consequently, a subset of $Z$ satisfies $\Pi_1$ iff
it satisfies~$\Pi_2$, which implies that $Z$ is minimal among the sets
satisfying $\Pi_1$ iff $Z$ is minimal among the sets satisfying $\Pi_2$.

\medskip\noindent{\sl Case 2:}
For some subset $Z'$ of $Z$, $Z'\models \Pi_1$ but
$Z'\not\models \Pi_2$.  Let $Z''$ be the intersection of all subsets
$X$ of $Z$ such that
\begin{enumerate}[(ii)]
\item[(i)]
$X\setminus Q=Z'\setminus Q$, and
\item[(ii)]
for every $q\in Q$, if $X \models \ii{Def}(q)$ then $q\in X$.
\end{enumerate}
We will establish several properties of $Z''$.  First,
\begin{equation}
\label{pr1}
Z''\subseteq Z'.
\end{equation}
Indeed, (i) holds for $Z'$ as $X$; since $Z'$ satisfies the program $\Pi_1$
that contains the rules $q\leftarrow \ii{Def}(q)$, (ii) holds for $Z'$ as well.
Consequently, $Z'$ is one of the sets~$X$
whose intersection we denoted by $Z''$, which implies (\ref{pr1}).

Second, $Z''$ satisfies conditions (i) and (ii) as $X$,
that is to say,
\begin{enumerate}[(ii$'$)]
\item[(i$'$)]
$Z''\setminus Q=Z'\setminus Q$, and
\item[(ii$'$)]
for every $q\in Q$, if $Z'' \models \ii{Def}(q)$ then $q\in Z''$.
\end{enumerate}
Property (i$'$) is a consequence of the fact that $Z''$ is the intersection of
a nonempty family of sets~$X$ satisfying (i).  To prove (ii$'$), take any
$q\in Q$ such that
$Z'' \models \ii{Def}(q)$. Each superset of $Z''$ satisfies $\ii{Def}(q)$
by Fact~\ref{fact2}. Each set $X$ that satisfies (i) and (ii) is
a superset of $Z''$, so that each of these sets~$X$ contains $q$ by (ii). As
$Z''$ is the intersection of these sets, $q\in Z''$.

By (i$'$), all literals from $Z'\setminus Z''$ belong to $Q$, and
consequently do not have regular occurences in the heads of the rules
of~$\Pi$.  Since $Z'\models\Pi$, we can conclude by Lemma~\ref{monotone}
that $Z''\models \Pi$.
By (ii$'$), $Z''$ satisfies the rules $q\leftarrow \ii{Def}(q)$.  Furthermore,
$Z''$ satisfies each rule $\ii{Def}(q)\leftarrow q$, because otherwise
$Z''\setminus \{q\}$ would have been a proper subset of $Z''$ that satisfies
conditions (i) and (ii) as $X$, which is impossible by the choice of $Z''$.
Consequently, $Z''\models \Pi_2$.  Since $Z'\not\models \Pi_2$, it follows
that $Z''$ is a proper subset of $Z'$.   Then $Z''$ is a proper subset of $Z$.
Since $Z$ has a proper subset satisfying $\Pi_2$, it is neither an answer
set for $\Pi_1$ nor an answer set for $\Pi_2$.
\end{proof}

We have proved the completion lemma for the case when $\Pi$ and the formulas
$\ii{Def}(q)$ do not contain negation as failure.  To prove the lemma in full
generality, apply this special case to program $\Pi^Z$ and the
formulas $\ii{Def}(q)^Z$.

\noindent\begin{proof}[Proof of the Lemma on Explicit Definitions]
Denote the set of rules $q\ar \ii{Def}(q)$ for all $q\in Q$ by $\Delta$.
The assertion of the lemma can be divided into two parts, and we will prove
them separately.

\medskip\noindent{\sl Claim 1: If $Z$ is an answer set for $\Pi\cup \Delta$
then $Z\setminus Q$ is an answer set for $\Pi$.}

\medskip
Consider first the case
when neither $\Pi$ nor $\Delta$ contains negation as failure.  Take an
answer set $Z$ for $\Pi\cup \Delta$ and a subset $Z'$ of $Z\setminus Q$.
Lemma~\ref{monotone} can be applied to program $\Delta$ and the subset
$(Z\cap Q)\cup Z'$ of $Z$, because $Z\setminus((Z\cap Q)\cup Z')$, as a part
of $Z\setminus Q$, does not contain literals occurring in the heads of the
rules of $\Delta$.  Consequently
\begin{equation}
\label{step1}
(Z\cap Q)\cup Z'\models \Delta.
\end{equation}
Since $Z$ is an answer set for $\Pi\cup \Delta$,
$$(Z\cap Q)\cup Z'\models \Pi\cup \Delta \hbox{ iff }
  (Z\cap Q) \cup Z'=Z   \hbox{ iff }Z'=Z\setminus Q.$$
Using~(\ref{step1}), we conclude:
$$(Z\cap Q)\cup Z'\models \Pi \hbox{ iff } Z'=Z\setminus Q.$$
Since no element of $Q$ occurs in $\Pi$, we can rewrite this as
$$Z'\models \Pi \hbox{ iff } Z'=Z\setminus Q.$$
Since $Z'$ here is an arbitrary subset of $Z\setminus Q$,
we proved that $Z\setminus Q$ is an answer set for $\Pi$.

To prove Claim 1 in the general case, consider an answer set $Z$ for
$\Pi\cup \Delta$.  It is an answer set for $\Pi^Z\cup \Delta^Z$ also.  By
the special case of Claim 1 proved above, $Z\setminus Q$ is an answer set for
$\Pi^Z$.  Since no element of $Q$ occurs in $\Pi$, $\Pi^{Z\setminus Q}=\Pi^Z$
(Lemma~\ref{samereduct}).  It follows that $Z\setminus Q$ is an answer
set for $\Pi^{Z\setminus Q}$, and consequently for~$\Pi$.

\medskip\noindent{\sl Claim 2: If $Z^*$ is an answer set for $\Pi$ then
there exists a unique answer set $Z$ for $\Pi\cup \Delta$ such that
$Z\setminus Q=Z^*$.}

Consider first the case
when neither $\Pi$ nor $\Delta$ contains negation as failure.
Let $Z^*$ be an answer set for $\Pi$.  Define
$$Z_0=Z^*\cup \{q\in Q\ :\ Z^*\models \ii{Def}(q)\}.$$
We will show that $Z_0$ is the only consistent set $Z$ of literals with
the properties from Claim 2.
Clearly $Z_0\setminus Q=Z^*$.  We will check now that
\begin{enumerate}[(iii)]
\item[(i)]
$Z_0$ satisfies $\Pi\cup \Delta$,
\item[(ii)]
no proper subset of $Z_0$ satisfies $\Pi\cup \Delta$, and
\item[(iii)]
every consistent set~$Z$ of literals
that satisfies $\Pi\cup \Delta$ and has the property $Z\setminus Q= Z^*$ is
a superset of $Z_0$.
\end{enumerate}

To show that $Z_0$ satisfies $\Pi$, observe that $Z^*$ satisfies $\Pi$ and
no element of $Q$ occurs in $\Pi$.
To show that $Z_0$ satisfies $\Delta$, assume that $Z_0\models \ii{Def}(q)$.
Since no element of $Q$ occurs in $\ii{Def}(q)$, it follows that
$Z^*\models \ii{Def}(q)$, so that $q\in Z_0$.  Assertion (i) is proved.

It is convenient to prove assertion (iii) next.
Take a consistent set~$Z$ of literals that satisfies $\Pi\cup \Delta$
and has the
property $Z\setminus Q= Z^*$.
First notice that
\begin{equation}
\label{step2}
Z_0\setminus Q
= Z^*=Z\setminus Q\subseteq Z.
\end{equation}
Take any $q\in Z_0\cap Q$.
Since $Z^*$ is disjoint from $Q$, $q$ belongs to the
second of the two sets whose union we denoted by $Z_0$, so that
$Z^*\models \ii{Def}(q)$.  Since $Z^*=Z\setminus Q$ and the elements of $Q$
do not occur in $\ii{Def}(q)$, it follows that $Z\models \ii{Def}(q)$.  In
view of the fact that $Z$ satisfies $\Delta$, we can conclude that $q\in Z$.
Since $q$ here is an arbitrary element of $Z_0\cap Q$, we proved that
$Z_0\cap Q\subseteq Z$.  In combination with (\ref{step2}), this fact shows
that $Z$ is a superset of $Z_0$.

To prove assertion (ii), assume that a proper subset $Z$ of $Z_0$ satisfies
$\Pi\cup\Delta$.
Since the elements of $Q$ do not occur in $\Pi$, it follows that
$Z\setminus Q$ satisfies $\Pi$.
On the other hand, $Z\setminus Q$ is a subset of $Z^*$.
Since $Z^*$ is an answer set
for $\Pi$, it follows that $Z\setminus Q$ cannot be a proper subset of $Z^*$.
Consequently $Z\setminus Q=Z^*$.  Then, by assertion (iii),
$Z$ is a superset of $Z_0$, which is impossible, by the choice of $Z$.

To prove Claim 2 in the general case, consider an answer set $Z^*$ for $\Pi$.
It is an answer set for $\Pi^{Z^*}$ also.  By the special case of Claim 2
proved above, there exists a unique answer set $Z$ for
$\Pi^{Z^*}\cup \Delta^{Z^*}$ such that $Z\setminus Q=Z^*$.
No element of $Q$ occurs in $\Pi$ or $\Delta$ in the scope of negation as
failure.  By Lemma~\ref{samereduct} it follows that
$\Pi^{Z^*}=\Pi^{Z}$ and $\Delta^{Z^*}=\Delta^{Z}$ for every $Z$  such that
$Z\setminus Q=Z^*$.  Consequently, there exists a unique answer set $Z$ for
$\Pi^{Z}\cup \Delta^{Z}$ such that $Z\setminus Q=Z^*$.  It follows that there
exists a unique answer set $Z$ for $\Pi\cup \Delta$ such that
$Z\setminus Q=Z^*$.
\end{proof}

\subsection{Proof of Theorem~\ref{th2}}

Let $\Omega$ be a program with weight constraints.  Consider the
subset~$\Delta$ of its nonnested translation~$[\Omega]^{nn}$ consisting
of the rules whose heads are atoms
from $Q_\Omega$.  The rules included in~$\Delta$ have the
forms~(\ref{lower1})--(\ref{choice-a}); they ``define''
the atoms in~$Q_\Omega$.  The rest of 
$[\Omega]^{nn}$ will be denoted
by~$\Pi$; the rules of~$\Pi$ have the
forms~(\ref{choice-b}) and~(\ref{constraint}).  The union of these two
programs is~$[\Omega]^{nn}$:
\begin{equation}
\label{opd}
[\Omega]^{nn}=\Pi\cup\Delta.
\end{equation}
The idea of the proof of Theorem~\ref{th2} is to transform $\Pi\cup\Delta$
into a program with the same answer sets so that $\Pi$ will turn into
$[\Omega]^{nd}$ and~$\Delta$ will turn into a set of explicit definitions
in the sense of Section~\ref{sec:lemmas}, and then use the lemma on
explicit definitions.

For every atom $q\in Q_\Omega$, define the formula $\ii{Def}(q)$ as follows:
$$
\begin{array}l
\ii{Def}(q_{\sno\ l})=\no\ l\\
\ii{Def}(q_{w\leq S})=
\begin{cases}
  \top,                                &\text{if $w\leq 0$,}\cr
  q_{w\leq S'};(c_m,q_{w-w_m\leq S'}), &\text{if $0<w\leq w_1+\dots+w_m$,}\cr
  \bot,                                &\text{otherwise}
\end{cases}\\
\ii{Def}(q_{w< S})=
\begin{cases}
  \top,                                &\text{if $w< 0$,}\cr
  q_{w< S'};(c_m,q_{w-w_m< S'}),       &\text{if $0\leq w< w_1+\dots+w_m$,}\cr
  \bot,                                &\text{otherwise}
\end{cases}\\
\end{array}
$$
\begin{lemma}
\label{th2l1}
Program $[\Omega]^{nn}$ has the same answer sets as
$$\Pi\cup\{q\leftrightarrow \ii{Def}(q)\; : \;q\in Q_\Omega\}.$$
\end{lemma}

\noindent\begin{proof}
>From the definitions of~$[\Omega]^{nn}$ and $Q_\Omega$ we conclude that
$\Delta$ consists of the following  rules:
\begin{enumerate}[$\bullet$]
\item[$\bullet$]
rule~(\ref{lower1}) for every atom of the form~$q_{w\leq S}$ in $Q_\Omega$
such that~$w\leq 0$;
\item[$\bullet$]
rules~(\ref{middle1}) for every atom of the form~$q_{w\leq S}\in Q_\Omega$
such that
$$0<w\leq w_1+\dots+w_m;$$
\item[$\bullet$]
rule~(\ref{lower2}) for every atom of the form~$q_{w<S}$ in $Q_\Omega$
such that~$w<0$;
\item[$\bullet$]
rules~(\ref{middle2}) for every atom of the form~$q_{w< S}$ in $Q_\Omega$
such that
$$0\leq w< w_1+\dots+w_m;$$
\item[$\bullet$]
rule~(\ref{choice-a}) for every atom of the form~$q_{\sno\ l}$ in $Q_\Omega$.
\end{enumerate}
Consequently~$\Delta$ is strongly equivalent to
$\{q\leftarrow \ii{Def}(q)\; : \;q\in Q_\Omega\}$.  Then,
by~(\ref{opd}), program~$[\Omega]^{nn}$ has the same answer sets as
$\Pi\cup \{q\leftarrow \ii{Def}(q)\; : \;q\in Q_\Omega\}$.  The assertion to be
proved follows by the completion lemma.
\end{proof}

\begin{lemma}
\label{recursive}
Let~$S$ be $\{c_1=w_1,\dots,c_m=w_m\}$.  In the logic of here-and-there,
$$
[w\leq S] \leftrightarrow
\begin{cases}
\top,                                  &\text{if $w\leq 0$},\cr
[w\leq S'];(c_m,[w-w_m\leq S']),       &\text{if $0<w\leq w_1+\cdots+w_m$,}\cr
\bot,                                  &\text{otherwise}.
\end{cases}
$$

$$
[w< S] \leftrightarrow
\begin{cases}
\top,                                  &\text{if $w< 0$},\cr
[w< S'];(c_m,[w-w_m< S']),             &\text{if $0\leq w<w_1+\cdots+w_m$,}\cr
\bot,                                  &\text{otherwise}.
\end{cases}
$$
\end{lemma}

\noindent\begin{proof}
Recall that $[w\leq S]$ is an expression of the form~(\ref{lowerold}),
which stands for a disjunction of conjunctions~(\ref{inlargeset}).
If $w\leq 0$ then the set after the : sign in~(\ref{lowerold}) has the empty
set as one of its elements, so that one of the disjunctive terms of this
formula is the empty conjunction $\top$. If $w> w_1+\cdots+w_m$ then the set
after the : sign in~(\ref{lowerold}) is empty, so that the formula is the empty
disjunction $\bot$.  Assume now that $0<w_1+\cdots+w_m\leq w$.
Let~$I$ stand for $\{1,\ldots,m\}$ and let~$I'$ be $\{1,\ldots,m-1\}$.
For any subset $J$ of $I$, by $\Sigma J$ we denote the sum $\sum_{i\in J} w_i$.
Then
$$
\begin{array}{rcl}
[w\leq S] &=&{\displaystyle
                   \bigscolon_{J\subseteq I\ :\ \Sigma J \geq w}
                  \big( \bigcomma_{i \in J} c_i\big)}\\ \\
              &\leftrightarrow&{\displaystyle
                   \bigscolon_{J\subseteq I'\ :\ \Sigma J \geq w}
                  \big( \bigcomma_{i \in J} c_i\big);
                   \bigscolon_{J\subseteq I\ :\ m\in J,\Sigma J \geq w}
                  \big( \bigcomma_{i \in J} c_i\big)}\\ \\
              &=&{\displaystyle
                   [w\leq S'];
                   \bigscolon_{J\subseteq I'\ :\ \Sigma J+w_m \geq w}
                  \big( \bigcomma_{i \in J\cup \{m\}} c_i\big)}\\ \\
              &\leftrightarrow&{\displaystyle
                   [w\leq S'];
                   \big(c_m, \bigscolon_{J\subseteq I'\ :\ \Sigma J \geq w-w_m}
                  \big( \bigcomma_{i \in J} c_i\big)\big)}\\ \\
              &=&{\displaystyle
                   [w\leq S'];(c_m,[(w-w_m)\leq S'])}.\\ \\
\end{array}
$$
The proof of the second equivalence is similar.
\end{proof}

\begin{lemma}
\label{th2l2}
Program
\begin{equation}
\label{eql2b}
\{q\leftrightarrow \ii{Def}(q)\; : \;q\in Q_\Omega\}
\end{equation}
is strongly equivalent to
\begin{equation}
\label{eql2}
\begin{array}l
\{q_{\sno\ l}\leftrightarrow\no\ l\  :\  q_{\sno\ l}\in Q_\Omega\}\cup\\
\{q_{w\leq S}\leftrightarrow[w\leq S]\  :\  q_{w\leq S}\in Q_\Omega\}\cup\\
\{q_{w< S}\leftrightarrow[w< S]\  :\  q_{w< S}\in Q_\Omega\}.
\end{array}
\end{equation}
\end{lemma}

\noindent\begin{proof}
The rules of (\ref{eql2}) can be obtained from the rules of (\ref{eql2b})
by replacing $\ii{Def}(q_{w\leq S})$ with $[w\leq S]$ for the atoms
$q_{w\leq S}$ in $Q_\Omega$, and
$\ii{Def}(q_{w<S})$ with $[w<S]$ for the atoms
$q_{w<S}$ in $Q_\Omega$.  Consequently, it is sufficient to show that, for
every atom of the form $q_{w\leq S}$ in $Q_\Omega$, the equivalences
\begin{equation}
\label{eql2a}
\ii{Def}(q_{w\leq S})\leftrightarrow[w\leq S]
\end{equation}
are derivable in the logic of here-and-there both from~(\ref{eql2b}) and
from~(\ref{eql2}), and similarly for atoms of the form $q_{w<S}$.
The proofs for atoms of both kinds are similar, and we will only
consider~$q_{w\leq S}$.  Let~$S$ be $\{c_1=w_1,\dots,c_m=w_m\}$.

The definition of $\ii{Def}(q_{w\leq S})$ and the statement of
Lemma~\ref{recursive} show that the right-hand side of~(\ref{eql2a}) is
equivalent to the result of replacing $q_{w\leq S'}$ in the
left-hand side with $[w\leq S']$, and $q_{w-w_m\leq S'}$ with $[w-w_m\leq S']$.
Since $q_{w\leq S'}$ and $q_{w-w_m\leq S'}$ belong to $Q_\Omega$,
this observation implies the derivability of~(\ref{eql2a}) from~(\ref{eql2}).

The derivability of~(\ref{eql2a}) from~(\ref{eql2b}) will be proved by strong
induction on~$m$.  If $w\leq 0$ or $w>w_1+\dots+w_m$ then, by the definition
of $\ii{Def}(q_{w\leq S})$ and by Lemma~\ref{recursive},~(\ref{eql2a}) is
provable in the logic of here-and-there.
Assume that $0<w\leq w_1+\dots+w_m$.  Then $q_{w\leq S'}$ and
$q_{w-w_m\leq S'}$ belong to $Q_\Omega$, and, by the induction hypothesis,
the equivalences
$$\ii{Def}(q_{w\leq S'}) \leftrightarrow [w\leq S']$$
and
$$\ii{Def}(q_{w-w_m\leq S'}) \leftrightarrow [w-w_m\leq S']$$
are derivable from~(\ref{eql2b}).  Consequently, the equivalences
$$q_{w\leq S'} \leftrightarrow [w\leq S']$$
and
$$q_{w-w_m\leq S'} \leftrightarrow [w-w_m\leq S']$$
are derivable from~(\ref{eql2b}) as well.  By Lemma~\ref{recursive},
this implies the derivability
of~(\ref{eql2a}).
\end{proof}

\begin{theorem2}
For any program $\Omega$ with weight constraints,
$Z\mapsto Z \setminus Q_\Omega$ is a 1--1 correspondence between the
answer sets for $[\Omega]^{nn}$ and the answer sets for $\Omega$.
\end{theorem2}

\noindent\begin{proof}
>From Lemmas~\ref{th2l1} and~\ref{th2l2} we see that $[\Omega]^{nn}$ has the
same answer sets as the union of $\Pi$ and (\ref{eql2}).  Furthermore, this
union is strongly equivalent to the union of~$[\Omega]^{nd}$ and~(\ref{eql2}).
Indeed, $\Pi$ consists of the rules
$$
\begin{array}l
      l \ar \no\ q_{\sno\ l},[C_1]^{nn},\dots,[C_n]^{nn},\\
      \bot \ar \no\ q_{L_0\leq S_0},[C_1]^{nn},\dots,[C_n]^{nn},\\
      \bot \ar q_{U_0<S_0},[C_1]^{nn},\dots,[C_n]^{nn}
\end{array}
$$
for every rule
$$
      L_0\leq S_0\leq U_0 \leftarrow C_1,\dots,C_n
$$
in $\Omega$ and every positive head element~$l$ of that rule;
$[\Omega]^{nd}$ consists of the rules
$$
\begin{array}l
l\ar \no\ \no\ l, [C_1],\ldots,[C_n],\\
\bot \ar \no\ [L_0\leq S_0,S_0\leq U_0], [C_1],\ldots,[C_n].
\end{array}
$$
It is easy to derive each of these two programs from the other program
and~(\ref{eql2}) in the logic of here-and-there.  Consequently,
$[\Omega]^{nn}$ has the same answer sets as the union of $[\Omega]^{nd}$
and (\ref{eql2}).  By the completion lemma, it follows that $[\Omega]^{nn}$
has the same answer sets as the union of $[\Omega]^{nd}$ and the
program
$$
\begin{array}l
\{q_{\sno\ l}\leftarrow\no\ l\  :\  q_{\sno\ l}\in Q_\Omega\}\cup\\
\{q_{w\leq S}\leftarrow[w\leq S]\  :\  q_{w\leq S}\in Q_\Omega\}\cup\\
\{q_{w< S}\leftarrow[w< S]\  :\  q_{w< S}\in Q_\Omega\}.
\end{array}
$$
The assertion of Theorem~\ref{th2} follows now by the lemma on explicit
definitions.
\end{proof}

\section{Conclusion}

The results of this paper show that weight constraints in the sense
of~\cite{nie00} can be viewed as shorthand for nested expressions.  Rules
with weight constraints can be equivalently written as sets of
nondisjunctive rules.  These rules can be further made nonnested, without
a significant increase in the size of the program, provided that auxiliary
atoms are allowed.  Moreover, when all weights are integers from a fixed
finite set, this translation leads to a program of
about the same size as the original program with weight constraints.
These facts, along with the
extension of the theory of tight programs proposed in~\cite{erd03}, have led
to the creation of the system \hbox{\sc cmodels}.  The ideas of this paper can
be also used to prove the strong equivalence of programs with weight
constraints.

\section*{Acknowledgments}

We are grateful to Selim Erdo\u gan for finding out several
inaccuracies, to Hudson Turner for his careful reading of a draft of this
paper and for many useful comments, and to the anonymous
referees for their suggestions.
This work was partially supported by National
Science Foundation under grant IIS-9732744 and by the Texas
Higher Education Coordinating Board under Grant 003658-0322-2001.

\bibliography{./bib}

\end{document}